\documentclass{article}

\usepackage{iclr2027_conference,times}

\usepackage[T1]{fontenc}
\usepackage[utf8]{inputenc}
\usepackage{microtype}

\usepackage{amsmath}
\usepackage{amssymb}
\usepackage{amsfonts}
\usepackage{amsthm}

\usepackage{graphicx}
\usepackage{wrapfig}
\usepackage{xcolor}
\usepackage{colortbl}
\usepackage{booktabs}
\usepackage{multirow}
\usepackage{makecell}
\usepackage{tabularx}
\usepackage{array}
\usepackage{caption}
\usepackage{tikz}
\usepackage{placeins}

\usepackage{hyperref}
\usepackage{url}

\theoremstyle{plain}

\theoremstyle{definition}

\theoremstyle{remark}

\title{WLA\textsuperscript{3}: World Latent Action Modeling\\ for Semantics, Dynamics, and Kinematics}

\author{Peidong~Liu$^{*,\dagger}$, Zhiyuan~Xiang$^*$, Mingyang~Li,
  Wenhao~Li, Jiale~Zhang, Jiahao~Sun, Jiawei~Li$^\dagger$ \\
  Joy Future Academy, JD Group \\
  \texttt{lpd19@tsinghua.org.cn     li-jw15@tsinghua.org.cn} \\
  \textnormal{\small $^*$Equal contribution. $^\dagger$Corresponding authors.}}

\iclrfinalcopy

\begin{document}

\maketitle
\fancyhead{}
\renewcommand{\headrulewidth}{0pt}

\begin{abstract}
Scaling generalist policy models with heterogeneous data is limited by the lack of
unified, low-noise action supervision. Human egocentric videos are abundant, but only
a small fraction comes with high-quality hand-action labels. Observed world transitions
offer a common source of action-related supervision across data sources. We introduce
\textbf{WLA$^3$} (\textbf{W}orld \textbf{L}atent \textbf{A}ction Modeling for
Semantics, Dynamics, and Kinematics), a unified generalist policy model framework
built around representations learned by a World Latent Action Model (WLAM). WLAM
first learns how multimodal world states change over a local interval, encoding
synchronized camera views and available embodiment-state changes into a compact local
latent action and a richer transition feature. Reconstruction from partial modalities and
consistency across overlapping windows encourage robust transition representations.
WLA$^3$ reuses them across semantics, dynamics, and kinematics: local latent actions support
action-sensitive physical-dynamics modeling, segment-level features directly supervise
the VLM through a Semantic Latent Aggregate (SLA), and an action expert jointly predicts
latent actions together with embodiment-specific robot controls. Human videos provide
scalable transition supervision, while robot trajectories ground the shared
representation in executable native controls. On LARYBench, the final 32D latent action
reaches 67.89\% average classification accuracy. WLA$^3$ achieves 81.9\% average
success across six real-robot tasks versus 66.2\% for $\pi_{0.5}$. Performance
improves as generalist policy model mid-training data scales, and human videos support
human-to-robot transfer.
Project page can be found at https://wla-3.github.io/.
\end{abstract}

\section{Introduction}
\label{sec:introduction}

\begin{figure*}[t]
    \centering
    \captionsetup{skip=4pt}
    \includegraphics[width=1\textwidth]{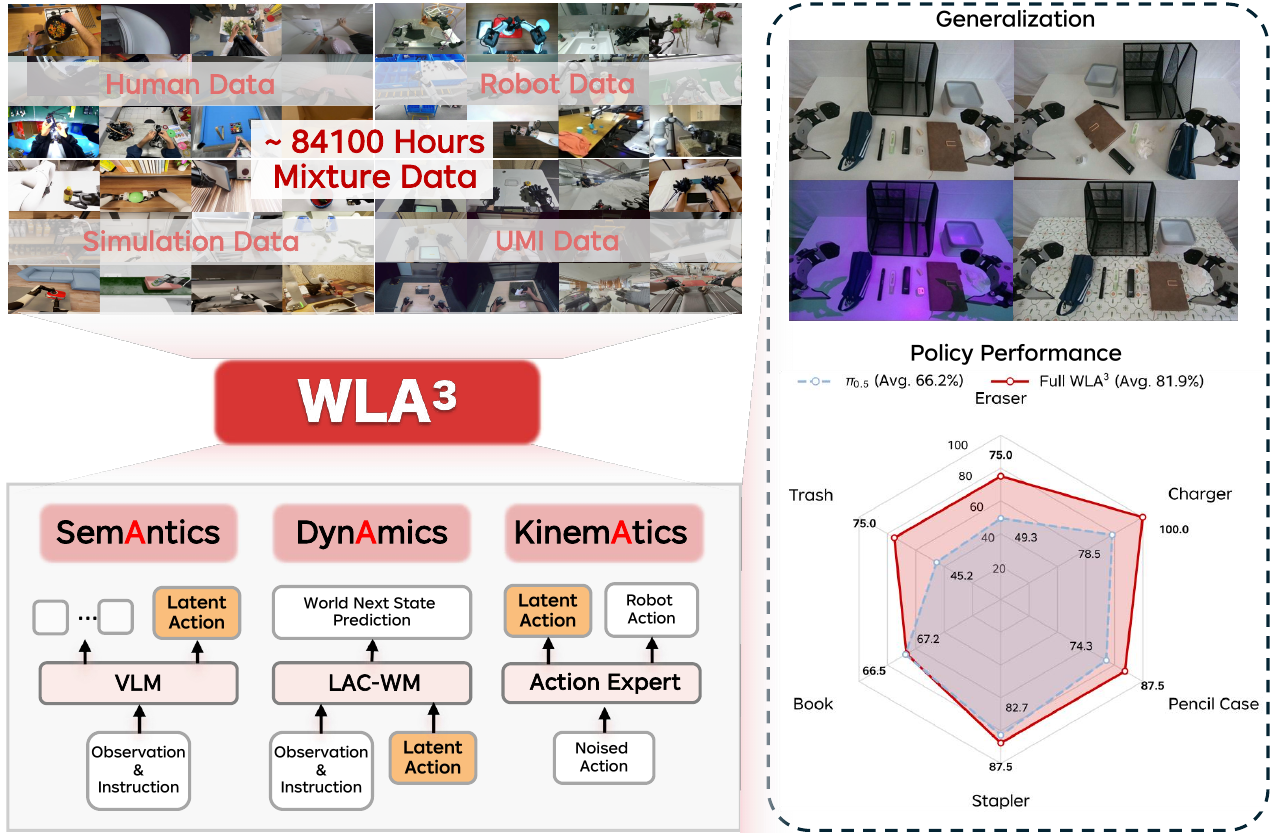}
    \caption{\textbf{Overview of WLA$^3$.} WLAM learns compact local latent actions
    and segment-level transition features from heterogeneous world-state transitions.
    These representations support latent action-conditioned dynamics learning, direct
    SLA supervision of the VLM, and joint latent--robot action generation.}
    \label{fig:overview}
\end{figure*}

Generalist policy models provide a general interface from visual observations and language instructions to robot control, yet scaling them with larger volumes of data from heterogeneous embodiments is limited by the lack of unified, low-noise action labels. High-quality robot trajectories remain scarce, and their native controls are tied to specific hardware configurations. In contrast, human videos offer abundant interactions, but existing 3D hand-pose estimates remain noisy and provide limited low-noise supervision. State transitions directly record the physical outcomes of interaction, including embodiment motion, contact changes, and evolving spatial geometry. They therefore provide a common basis for representing action across data sources, while native controls preserve the embodiment-specific information needed for execution.

Latent Action Models (LAMs) exploit this insight by compressing state transitions into latent actions~\citep{ye2024latent,bu2025univla}. However, existing LAMs face two critical limitations. First, they typically infer latents from single-view image pairs, where occlusion and perspective projection render distinct physical interactions visually ambiguous. Recent methods improve physical grounding through contrastive learning in CLAP~\citep{zhang2026clap}, state decoding in villa-X~\citep{chen2025villa}, or auxiliary viewpoints for primary-view learning in MVP-LAM~\citep{lee2026mvplam}; these methods still construct latent actions primarily from primary-view transitions and overlook the additional world-state information that other views and states could contribute. Second, prior works typically tailor latent actions to isolated downstream roles rather than treating them as a shared interface across the full generalist policy model stack. Different components require transition knowledge at distinct temporal scales: physical dynamics and low-level control benefit from compact local targets, while a condensed segment-level target provides a more direct gradient for grounding the VLM's instruction-conditioned representation in the physical transition of a policy segment.

To address these challenges, as Figure~\ref{fig:overview} shows, we introduce \textbf{WLA$^3$} (\textbf{W}orld \textbf{L}atent \textbf{A}ction Modeling for Semantics, Dynamics, and Kinematics), a unified generalist policy model framework that connects latent action modeling, world modeling, and policy learning. At its foundation, a World Latent Action Model (WLAM) introduces world-state modeling into latent action learning by jointly encoding synchronized camera views and, when available, proprioceptive states. Cross-modal dropout prevents reliance on any single modality by requiring partial transition evidence to reconstruct the recorded endpoint modalities. A cosine-distance objective further encourages consistent latent-action directions across overlapping temporal windows~\citep{tang2026alam,wang2026temporalstraightening}.

After WLAM training, WLA$^3$ uses its transition representations at different granularities through LAC-WM pre-training, generalist policy model mid-training, and generalist policy model post-training. A Latent Action-Conditioned World Model (LAC-WM) conditions on compact continuous latents to capture action-sensitive physical dynamics, while a Semantic Latent Aggregate (SLA) mean-pools WLAM features over a policy segment and directly supervises the VLM to predict this segment-level transition feature from current observations and instructions. A Latent-Action and Robot-Action (LARA) expert jointly predicts latent actions and executable robot actions. The latent branch retains consistent transition semantics across training stages and embodiments, allowing heterogeneous robot data to share a common latent-action target. Human videos provide visual transition-derived latent supervision, while robot trajectories enrich the same representation with proprioceptive transitions and ground it in executable native controls, enabling scalable human-to-robot transfer.

We evaluate policy performance through real-robot manipulation success rates, data
scaling, and human-to-robot alignment, and evaluate the learned latent action
representations on LARYBench. The final 32D latent action reaches 67.89\% average
classification accuracy on LARYBench, while WLA$^3$ achieves 81.9\% average success
across six real-robot tasks versus 66.2\% for $\pi_{0.5}$; success rises from 57.6\%
to 81.9\% as generalist policy model mid-training data scales from 5K to 20K hours. Our main contributions are:
\begin{itemize}
    \item \textbf{World Latent Action Modeling.} WLAM introduces world-state modeling
    into latent action learning by jointly encoding primary and hand-view observations
    with available proprioceptive states. Cross-modal dropout and a cosine-distance
    objective over overlapping temporal windows regularize the latent action space.

    \item \textbf{Generalist Policy Design.} Motivated by the distinct ways transition
    knowledge supports semantics, dynamics, and kinematics, we organize policy
    learning around LAC-WM for latent action-conditioned physical dynamics,
    Semantic Latent Aggregate (SLA) supervision that condenses transition evidence
    into a direct training target for the VLM, and LARA for joint latent and robot
    action generation.

    \item \textbf{Extensive Evaluation.} We evaluate real-robot manipulation success
    rates, scaling with heterogeneous data, and human-to-robot alignment, and further
    assess the learned latent action representations on LARYBench.
\end{itemize}

\section{Related Work}
\label{sec:related_work}

\subsection{Latent Action Modeling}

Latent Action Models (LAMs) infer action representations from observation
transitions, providing supervision from videos and trajectories without requiring
a shared native control space. Genie~\citep{bruce2024genie} learns discrete latent
actions for controllable video generation, while LAPA~\citep{ye2024latent} and
UniVLA~\citep{bu2025univla} use video-derived or task-centric latent actions for
generalist policy model pre-training. CLAM~\citep{liang2025clam} and CoMo~\citep{yang2025como} instead
learn continuous latent action or motion representations, and
Motus~\citep{bi2025motus} couples latent actions with world modeling. Other works
improve the grounding or organization of latent actions:
villa-X~\citep{chen2025villa} introduces robot-state decoding, CLAP~\citep{zhang2026clap}
and ConLA~\citep{dai2026conla} use contrastive alignment, and
MVP-LAM~\citep{lee2026mvplam} uses cross-viewpoint reconstruction to encourage
action-centric representations. HiLAM~\citep{kim2026hilam} introduces temporal
hierarchy, RotVLA~\citep{li2026rotvla} imposes rotational structure,
DiLA~\citep{zhang2026dila} disentangles latent factors, and
ALAM~\citep{tang2026alam} promotes algebraic consistency. Building on these advances,
WLAM jointly encodes synchronized camera views and
available proprioceptive states, applies cross-modal dropout, and regularizes
overlapping temporal windows with a cosine objective. It produces both a compact
continuous latent action for local physical dynamics and control and a
pre-bottleneck transition feature that can be condensed into a direct semantic
training target for the VLM.

\subsection{Generalist Policy Learning}

Recent generalist policies adapt pretrained vision-language models (VLMs) into
generalist policy models for language-conditioned control. These methods
differ in action representation, embodiment alignment, and predictive
supervision. Open-world generalization is pursued through heterogeneous
co-training in $\pi_{0.5}$~\citep{intelligence2025pi_}, embodied reasoning in
Gemini Robotics~\citep{geminirobotics2025}, and cross-embodiment transfer in
X-VLA~\citep{zheng2025xvla} and Gemini Robotics
1.5~\citep{geminirobotics15_2025}. LingBot-VLA~\citep{wu2026lingbotvla} and Galaxea
G0.5~\citep{galaxea2026g05} emphasize scalable pre-training and autoregressive
actions, while Qwen-RobotManip~\citep{yuan2026qwenrobotmanip} and
Being-H0.5~\citep{luo2026beingh05} align representations across embodiments.
Hy-Embodied-0.5-VLA~\citep{zhang2026hyembodied} and
$\pi_{0.7}$~\citep{physicalintelligence2026pi07} target deployment and steering,
whereas InternVLA-A1~\citep{cai2026internvla} and
Wall-OSS-0.5~\citep{yu2026walloss05technicalreport} unify understanding,
generation, and action.

A parallel direction equips generalist policies with predictive world models.
IRASim~\citep{zhu2025irasim} and Unified World
Models~\citep{zhu2025unifiedworldmodels} learn action-conditioned visual
dynamics, while UP-VLA~\citep{zhang2025upvla} and
WorldVLA~\citep{cen2025worldvla} connect future prediction with policy learning.
VideoVLA~\citep{shen2025videovla} and FLARE~\citep{zheng2025flare} transfer
generative or implicit dynamics priors to control; Motus~\citep{bi2025motus} and
LingBot-VA~\citep{li2026lingbotva} more directly unify video, language, and
action prediction. LDA-1B~\citep{lyu2026lda} and
DreamDojo~\mbox{\citep{gao2026dreamdojo}} scale dynamics learning across heterogeneous
embodied or human video data, whereas VLA-JEPA~\citep{sun2026vlajepa} and
FRAPPE~\citep{zhao2026frappe} align policy representations with future features.
JoyAI-RA 0.5~\citep{joyairateam2026joyaira05} combines latent-action-conditioned
dynamics with dual action alignment for heterogeneous policy learning. WLA$^3$
uses multimodal world-transition representations at multiple granularities
across semantics, physical dynamics, and control.

\subsection{Cross-Embodiment Learning}

Recent cross-embodiment learning scales heterogeneous robot and human data through shared motion representations and flexible policy architectures. Gemini Robotics 1.5~\citep{geminirobotics15_2025} introduces Motion Transfer for zero-shot skill transfer across robot embodiments, while $\pi_{0.7}$~\citep{physicalintelligence2026pi07} demonstrates out-of-the-box control on a new embodiment through diverse context-conditioned training. Qwen-RobotManip~\citep{yuan2026qwenrobotmanip} uses a canonical state--action space, camera-frame end-effector motions, and human-to-robot synthesis across multiple embodiments, and GR00T N1.7~\citep{nvidia2026grootn17} combines a relative end-effector interface with large-scale egocentric-video pre-training. Being-H0.5~\citep{luo2026beingh05} uses semantically aligned action slots and shared and embodiment-specific flow experts, whereas Being-H0.7~\citep{luo2026beingh07} learns future-aware latent reasoning from egocentric videos.

WLA$^3$ uses a complementary transition-centered representation: multimodal local latent actions provide a common motion target across data sources, while robot controls retain their embodiment-specific coordinates. The same WLAM representations support the VLM, world model, and policy at different temporal scales.

\section{WLA\textsuperscript{3}: World Latent Action Modeling}
\label{sec:lam}

\subsection{World Latent Action Model}
\label{sec:latent_action_model}

\begin{figure*}[!htbp]
    \centering
    \includegraphics[width=1\textwidth]{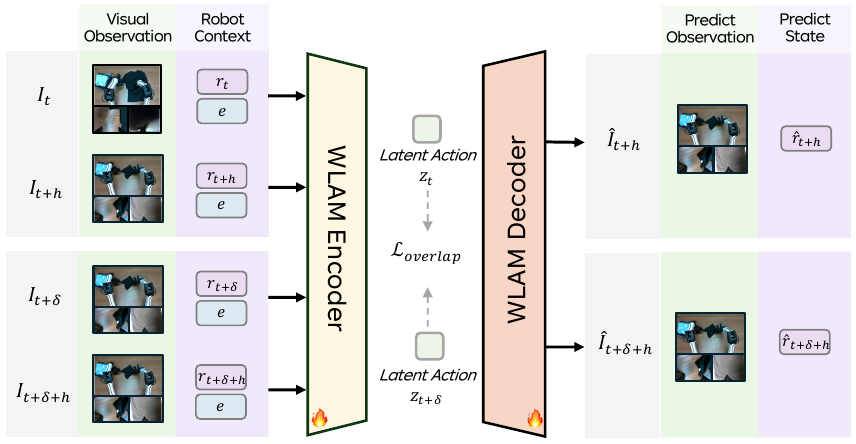}
    \caption{\textbf{World latent action modeling.} WLAM jointly encodes camera transitions and, when available, embodiment-state transitions into a compact latent action, then reconstructs the endpoint observations. Cross-modal transition dropout promotes a shared representation of physical transitions across views and embodiment states. For overlap consistency, two \(h\)-step transition windows start \(\delta\) steps apart, where \(0<\delta<h\), and thus overlap for \(h-\delta\) steps.}
    \label{fig:lam}
\end{figure*}

WLAM is trained on the heterogeneous mixture in Section~\ref{sec:dataset}, which
contains human videos and state-bearing robot, simulation, and UMI trajectories. At
time $t$, $I_t$ denotes the available camera observations, $r_t$ the recorded
embodiment state, and $e$ the embodiment profile. We write the world state as
$s_t=(I_t,r_t,e)$ and omit modalities that are not recorded. All four data families
contribute to a shared transition representation. The WLAM architecture is shown in Figure~\ref{fig:lam}.

\paragraph{World-Transition Encoding.}
\label{sec:lam_segment_representation}
Given the endpoint states $s_t$ and $s_{t+h}$, the transition encoder jointly
processes their visual observations and available embodiment states, conditioned on
the embodiment profile $e$. We take the output of a learned $[\mathrm{CLS}]$ token as
the 1024D transition feature
\begin{equation}
    \mathbf{g}_{t,h}=\left[E^{\mathrm{trans}}_\phi(s_t,s_{t+h})\right]_{[\mathrm{CLS}]}\in\mathbb{R}^{1024}.
\end{equation}
The feature parameterizes a Gaussian posterior over a compact continuous latent
action~\citep{kingma2014autoencoding}:
\begin{equation}
    \mathbf{z}_{t,h}\sim q_\phi(\,\cdot\mid \mathbf{g}_{t,h})
    \equiv q_\phi(\,\cdot\mid s_t,s_{t+h}),
    \qquad \mathbf{z}_{t,h}\in\mathbb{R}^{d_z},\quad d_z=32.
\end{equation}
We use posterior samples for reconstruction and the posterior mean
$\bar{\mathbf{z}}_{t,h}=\mathbb{E}_{q_\phi(\,\cdot\mid
s_t,s_{t+h})}[\mathbf{z}]$ as the deterministic target for consistency
regularization and downstream training. For a segment spanning $t$ to $t+T$, we
mean-pool the local transition features:
\begin{equation}
    \mathbf{c}_t^{\mathrm{seg}}=\frac{1}{T-h+1}\sum_{k=0}^{T-h}\mathbf{g}_{t+k,h}\in\mathbb{R}^{1024}.
    \label{eq:segment_transition_pooling}
\end{equation}
During WLAM pre-training, the endpoint offset $h$ may exceed one step.
For offline target construction in subsequent stages, we set $h=1$ and extract consecutive transitions over one control interval. Here, $T$ denotes the segment horizon and $h$ is the endpoint offset of each local
transition. The deterministic latent action $\bar{\mathbf{z}}_{t,h}$ provides a local target
for latent-action prediction and latent-action-conditioned dynamics, while
$\mathbf{c}_t^{\mathrm{seg}}$ provides the segment-level SLA target for direct VLM
supervision. Section~\ref{sec:policy_model} describes how the policy uses both
representations. 
Posterior parameterization and input tokenization are detailed in Appendix~\ref{app:wlam_details}.

\paragraph{Cross-Modal World-State Reconstruction.}
WLAM applies cross-modal transition dropout so that the latent action can be inferred
from different subsets of the recorded transition. During training, we randomly drop
available camera or embodiment-state streams while retaining at least one stream. The
same mask is applied at $t$ and $t+h$, preventing a missing input from appearing as a
physical change. We denote the masked endpoints by $\widetilde{s}_t$ and
$\widetilde{s}_{t+h}$. The decoder receives the recorded initial state and
reconstructs all available endpoint modalities. Modalities absent from a source are
omitted from both the encoder input and the corresponding reconstruction terms.
Complete transitions are used when extracting offline targets. 
Appendix~\ref{app:wlam_details} specifies the masking and missing-modality handling.

\paragraph{Regularization and Training Objective.}
Temporally shifted windows often contain largely overlapping physical changes. During WLAM pre-training, overlap consistency is applied when
$h>1$. For two
length-$h$ windows separated by $0<\delta<h$, we minimize the cosine distance between
their posterior means~\citep{tang2026alam,wang2026temporalstraightening}:
\begin{equation}
    \mathcal{L}_{\mathrm{overlap}}=\mathbb{E}_{t,h,\delta:\,0<\delta<h}\!\left[
    1-\frac{\bar{\mathbf{z}}_{t,h}^{\top}\bar{\mathbf{z}}_{t+\delta,h}}
    {\lVert \bar{\mathbf{z}}_{t,h}\rVert_2
    \lVert \bar{\mathbf{z}}_{t+\delta,h}\rVert_2+\varepsilon_{\mathrm{cos}}}\right],
    \qquad \varepsilon_{\mathrm{cos}}>0.
\end{equation}

This regularizer favors locally consistent latent directions without directly
constraining transition magnitude. The reconstruction term $\mathcal{L}_{\mathrm{rec}}$
reconstructs the observed scene and, when available, embodiment-state changes. Together
with overlap consistency and the variational regularizer, the WLAM objective is
\begin{equation}
    \mathcal{L}_{\mathrm{WLAM}}
    =\mathcal{L}_{\mathrm{rec}}
    +\lambda_{\mathrm{overlap}}\mathcal{L}_{\mathrm{overlap}}
    +\beta D_{\mathrm{KL}}\!\left(
    q_\phi(\,\cdot\mid\widetilde{s}_t,\widetilde{s}_{t+h})
    \,\|\,\mathcal{N}(0,I_{d_z})\right).
\end{equation}
The decomposition of $\mathcal{L}_{\mathrm{rec}}$ and the temporal-window construction
are provided in Appendix~\ref{app:wlam_details}.

\subsection{Generalist Policy Model}
\label{sec:policy_model}

The generalist policy model uses WLAM representations at two temporal scales. Compact
latent actions describe the physical change over one control interval, while mean-pooled
transition features summarize a 30-step policy segment as a Semantic Latent Aggregate
(SLA) target. Together, they connect segment-level semantics and local dynamics to
executable robot control.

Before generalist policy model post-training, training proceeds through LAC-WM
pre-training and generalist policy model mid-training. LAC-WM pre-training uses
WLAM-provided local latent actions to condition the Latent Action-Conditioned World
Model (LAC-WM)~\citep{joyairateam2026joyaira05}. Generalist policy model mid-training
uses SLA to supervise the VLM directly and trains a Latent-Action and Robot-Action
(LARA) expert to jointly predict latent-action and robot-action chunks. LARA conditions
on semantic features from the VLM and dynamics features from the frozen LAC-WM. The
resulting policy is then adapted during generalist policy model post-training.

\begin{figure*}[t]
    \centering
    \includegraphics[width=\textwidth]{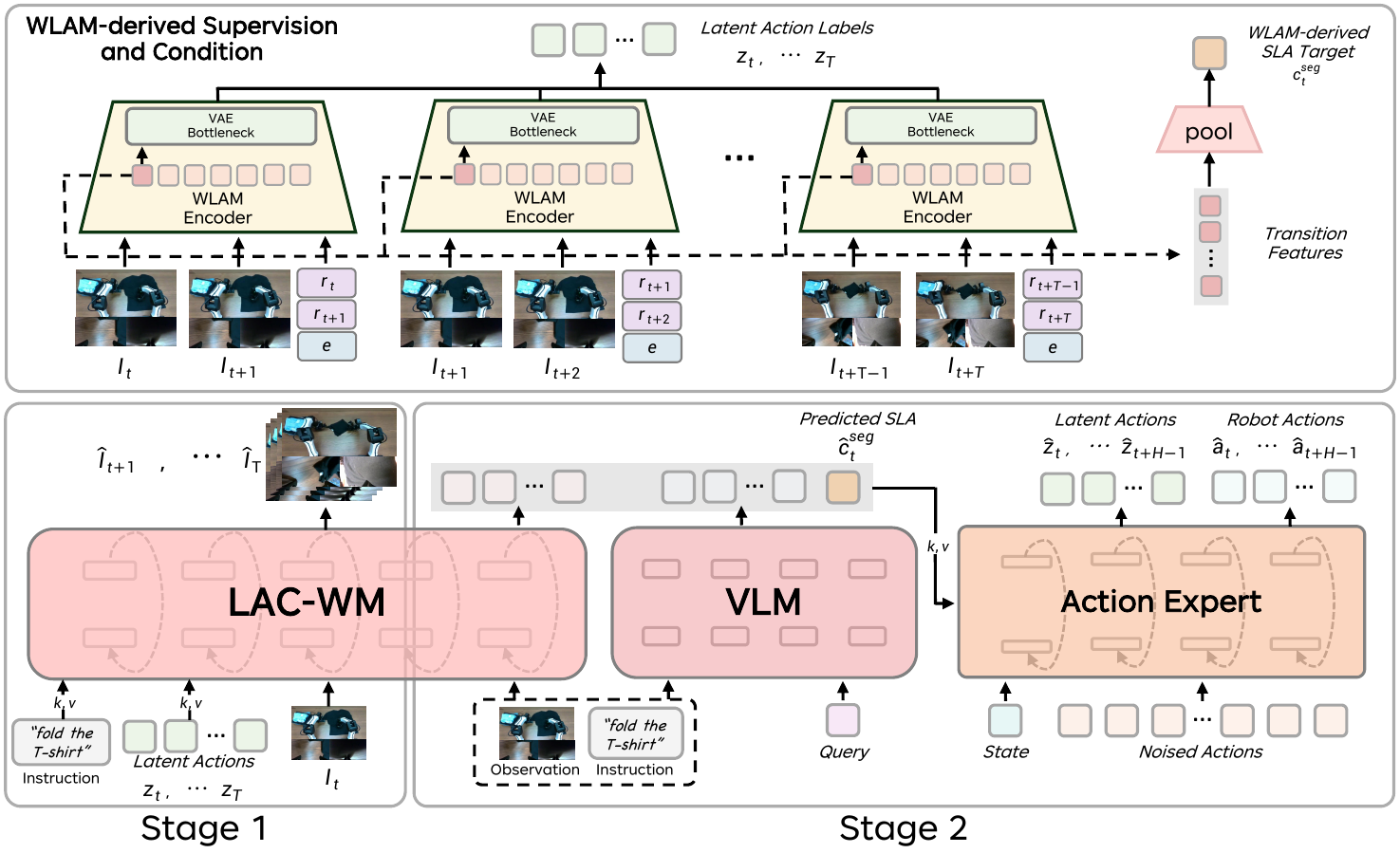}
    \caption{\textbf{LAC-WM pre-training and generalist policy model mid-training.}
    During LAC-WM pre-training, the frozen WLAM extracts latent actions from
    world-state transitions, which condition LAC-WM with the current
    observation and instruction.
    During generalist policy model mid-training, pooled WLAM features form the SLA target
    for direct VLM supervision, while local latent actions supervise LARA. The action expert
    combines VLM and frozen LAC-WM features to predict latent and robot actions jointly.
    Fire and snowflake symbols denote trainable and frozen modules.}
    \label{fig:wm_policy_training}
\end{figure*}

\paragraph{Dynamics: Latent Action-Conditioned World Modeling.}
As illustrated by LAC-WM pre-training in Figure~\ref{fig:wm_policy_training}, the current
observation and instruction alone may correspond to multiple possible physical
transitions. The frozen WLAM annotates each sequence with one compact latent action per
control interval; conditioning future prediction on these latents helps LAC-WM distinguish
the physical effects of different interactions. Given the current visual feature
$\mathbf{f}_t=E_{\mathrm{vis}}(I_t)$, task instruction $\ell$, and an $H_v$-step
latent-action chunk
$\mathbf{Z}_t=(\bar{\mathbf{z}}_{t,h},\ldots,\bar{\mathbf{z}}_{t+H_v-1,h})$, let
$\boldsymbol{\kappa}_t$ contain the retained subset of the language and latent-action
conditions. Let $\mathbf{y}=\mathbf{f}_{t+1:t+H_v}$ be the target future visual feature,
$\boldsymbol{\epsilon}$ be standard Gaussian noise of the same shape, and
$\tau\sim\mathcal{U}(0,1)$. Conditional Flow Matching uses the interpolation and target
velocity
\begin{equation}
    \mathbf{x}_{\tau}=(1-\tau)\boldsymbol{\epsilon}+\tau\mathbf{y},
    \qquad \mathbf{v}^{\star}_{\tau}=\mathbf{y}-\boldsymbol{\epsilon}.
    \label{eq:policy_lacwm_flow_path}
\end{equation}

For the noised future feature $\mathbf{x}_{\tau}$, LAC-WM predicts the velocity field
\begin{equation}
    \widehat{\mathbf{v}}_{\tau}
    =u_{\psi}\!\left(\mathbf{x}_{\tau},\tau;
    \mathbf{f}_t,\boldsymbol{\kappa}_t\right),
    \qquad \boldsymbol{\kappa}_t\subseteq\{\ell,\mathbf{Z}_t\}.
    \label{eq:policy_lacwm_velocity}
\end{equation}
Latent actions specify local transition cues, language specifies the task, and their joint
condition connects an instruction to its physical dynamics. Language and latent actions
are dropped independently during pre-training, exposing joint, language-only, latent-only,
and observation-only conditioning modes. In particular, language-only samples train the
observation--language pathway used during generalist policy model mid-training.

Following DreamDojo~\citep{gao2026dreamdojo}, LAC-WM is trained with velocity regression
and temporal-difference regularization:
\begin{equation}
    \mathcal{L}^{\mathrm{pre}}_{\mathrm{WM}}
    =\mathbb{E}\!\left[
    \|\widehat{\mathbf{v}}_{\tau}-\mathbf{v}^{\star}_{\tau}\|_2^2
    +\lambda_{\mathrm{temporal}}
    \|\Delta_{\mathrm{pos}}\widehat{\mathbf{v}}_{\tau}
    -\Delta_{\mathrm{pos}}\mathbf{v}^{\star}_{\tau}\|_2^2\right],
    \label{eq:policy_lacwm_objective}
\end{equation}
where, for position $i$, $\Delta_{\mathrm{pos}}\mathbf{v}_{\tau,i}
=\mathbf{v}_{\tau,i+1}-\mathbf{v}_{\tau,i}$. The first term trains conditional
future-feature prediction, while the second encourages neighboring velocity predictions
to vary coherently.

After LAC-WM pre-training, LAC-WM is frozen and provides dynamics-aware context from the
current visual feature and instruction during generalist policy model mid-training. We
denote its causal first-frame features by $\mathbf{H}_t^{\mathrm{wm}}$. The VLM receives
the current observation directly; future observations and WLAM outputs are used to
construct offline targets.

\paragraph{Semantics: Semantic Latent Aggregate Supervision.} SLA directly supervises
the VLM at the segment level. WLAM transition features are mean-pooled over a 30-step
policy segment to form the 1024D SLA target
$\mathbf{c}_t^{\mathrm{seg}}$ in Eq.~\ref{eq:segment_transition_pooling}. During
generalist policy model mid-training, a learned semantic query is appended to the VLM
input. Given the current observation and instruction, the query state and its projected
SLA prediction are
\begin{equation}
    \mathbf{H}_t^{\mathrm{vlm}}=V_{\xi}(I_t,\ell,\mathbf{q}_{\mathrm{sem}}),
    \qquad
    \widehat{\mathbf{c}}_t^{\mathrm{seg}}=
    P_{\mathrm{sem}}\!\left(\mathbf{H}_t^{\mathrm{vlm}}
    [\mathbf{q}_{\mathrm{sem}}]\right).
    \label{eq:policy_segment_prediction}
\end{equation}
The projection head maps the semantic-query state to the SLA dimension. Direct supervision
of this prediction transfers segment-level transition evidence into the VLM representation:
\begin{equation}
    \mathcal{L}_{\mathrm{sem}}=
    \left\|\widehat{\mathbf{c}}_t^{\mathrm{seg}}
    -\mathbf{c}_t^{\mathrm{seg}}\right\|_2^2.
    \label{eq:policy_segment_loss_main}
\end{equation}
$\mathcal{L}_{\mathrm{sem}}$ constitutes the semantic-supervision term in the
overall mid-training objective defined in
Eq.~\ref{eq:policy_objective_main}.

\paragraph{Kinematics: Latent Action-Guided Robot Control.} Native robot actions differ
across embodiments in dimension and control meaning, and their coordinates remain
embodiment-specific. Latent actions provide a shared local transition target, while a
separate robot-action branch grounds this representation in executable controls for each
embodiment.

For embodiment $e$, the policy uses separate fixed 32-dimensional interfaces
for state conditioning and robot-action prediction. The native state is
zero-padded to form the policy condition $\mathbf{r}_t^{\mathrm{pol}}$, while
human-video samples use $\mathbf{r}_t^{\mathrm{pol}}=\mathbf{0}_{32}$. The
native action has dimension $d_e\leq32$ and occupies the leading coordinates
of the robot-action interface; the remaining coordinates are filled with
zeros and masked out of the loss:
\begin{equation}
    \mathbf{a}_k^{\mathrm{rob}}
    =\left[\mathbf{a}_k^{(e),\mathrm{native}};\mathbf{0}_{32-d_e}\right]
    \in\mathbb{R}^{32},\qquad
    \mathbf{m}_k^{\mathrm{rob}}
    =\left[\mathbf{1}_{d_e};\mathbf{0}_{32-d_e}\right].
    \label{eq:policy_robot_action_interface}
\end{equation}
The mask selects valid coordinates for the robot-action loss, which are interpreted in the
target embodiment's native format at inference.

The SLA segment and LARA action chunk cover the same 30 control steps, so
$T=H_{\mathrm{act}}=30$. At each step, the deterministic local transition target and
robot action form a joint vector:
\begin{equation}
    \mathbf{x}_k
    =\left[\bar{\mathbf{z}}_{k,h}\mathbin{\Vert}\mathbf{a}_k^{\mathrm{rob}}\right]
    \in\mathbb{R}^{64},\qquad
    \mathbf{X}_t=(\mathbf{x}_t,\ldots,\mathbf{x}_{t+H_{\mathrm{act}}-1}).
    \label{eq:policy_joint_chunk}
\end{equation}
Here $\bar{\mathbf{z}}_{k,h}$ represents the local transition from step $k$ to $k+h$.
Before action prediction, frozen LAC-WM features are projected to the VLM hidden
dimension and concatenated with the VLM context:
\begin{equation}
    \mathbf{H}_t^{\mathrm{pol}}
    =\left[\mathbf{H}_t^{\mathrm{vlm}};
    \operatorname{LN}\!\left(P_{\mathrm{wm}}
    \!\left(\mathbf{H}_t^{\mathrm{wm}}\right)\right)\right].
    \label{eq:policy_context_fusion_main}
\end{equation}
We model the joint action chunk with conditional Flow Matching~\citep{lipman2023flowmatching}.
The time-conditioned Perceiver action expert~\citep{jaegle2021perceiver} receives Gaussian
noise $\boldsymbol{\epsilon}\in\mathbb{R}^{H_{\mathrm{act}}\times64}$, a noised
joint-action chunk at $\tau\sim\mathcal{U}(0,1)$, the state condition, and the fused
policy context, and predicts two output branches:
\begin{equation}
\begin{aligned}
    \widetilde{\mathbf{X}}_{t,\tau}&=(1-\tau)\boldsymbol{\epsilon}+\tau\mathbf{X}_t,
    &\mathbf{V}_{t,\tau}^{\star}&=\mathbf{X}_t-\boldsymbol{\epsilon},\\
    \widehat{\mathbf{V}}_{t,\tau}
    &=F_{\eta}\!\left(\widetilde{\mathbf{X}}_{t,\tau},
    \mathbf{r}_t^{\mathrm{pol}},\mathbf{H}_t^{\mathrm{pol}},\tau\right)
    =\left[\widehat{\mathbf{V}}_{t,\tau}^{\mathrm{lat}}\mathbin{\Vert}
    \widehat{\mathbf{V}}_{t,\tau}^{\mathrm{rob}}\right].
\end{aligned}
    \label{eq:joint_action_prediction}
\end{equation}
Numerical integration of this velocity field produces the predicted latent-action and
robot-action chunk. The two branches use masked Flow Matching losses, and each branch is
normalized by its own exponential moving average so that their relative scales remain
balanced during training. Appendix~\ref{app:policy_details} gives the masked branch
losses and EMA normalization explicitly.

Latent-only trajectories contribute to the latent branch, while trajectories with both
targets, including UMI data, contribute to both branches. The generalist policy model
mid-training objective combines
the latent-action, robot-action, and SLA losses:
\begin{equation}
    \mathcal{L}_{\mathrm{policy}}
    =\lambda_{\mathrm{lat}}\widetilde{\mathcal{L}}_{\mathrm{lat}}
    +\lambda_{\mathrm{rob}}\widetilde{\mathcal{L}}_{\mathrm{rob}}
    +\lambda_{\mathrm{sem}}\mathcal{L}_{\mathrm{sem}}.
    \label{eq:policy_objective_main}
\end{equation}
Generalist policy model post-training retains the latent-action and robot-action prediction
branches while removing the semantic query, SLA prediction head, and semantic loss. At
deployment, numerical integration produces both branches, but only the valid coordinates
of the robot-action branch are executed in the target embodiment's native action format.

\FloatBarrier

\section{Datasets}
\label{sec:dataset}

\definecolor{datasetstageWLAM}{HTML}{F2C99F}
\definecolor{datasetstageLACWM}{HTML}{A9D9EA}
\definecolor{datasetstageGPM}{HTML}{B9DDB5}
\newcommand{\datasetstagemarker}[2]{%
    \tikz[baseline=(datasetstage.base)]{%
        \node[circle, fill=#1, minimum size=1.45em, inner sep=0pt,
        font=\bfseries\scriptsize] (datasetstage) {#2};}}
\newcommand{\datasetstageWLAMmarker}{\datasetstagemarker{datasetstageWLAM}{W}}
\newcommand{\datasetstageLACWMmarker}{\datasetstagemarker{datasetstageLACWM}{L}}
\newcommand{\datasetstageGPMmarker}{\datasetstagemarker{datasetstageGPM}{G}}
\newcommand{\datasetstageWLAMLACWM}{\datasetstageWLAMmarker\hspace{0.18em}%
    \datasetstageLACWMmarker}
\newcommand{\datasetstageWLAMGPM}{\datasetstageWLAMmarker\hspace{0.18em}%
    \datasetstageGPMmarker}
\newcommand{\datasetstageall}{\datasetstageWLAMmarker\hspace{0.18em}%
    \datasetstageLACWMmarker\hspace{0.18em}\datasetstageGPMmarker}
\newcommand{\datasetname}[1]{#1}

Figure~\ref{fig:dataset_overview} summarizes the corpus composition,
representative observations, and key data attributes, while
Table~\ref{tab:lam_data} provides the dataset-level breakdown
and indicates which training stages use each source.

\begin{center}
    \includegraphics[width=1\textwidth]{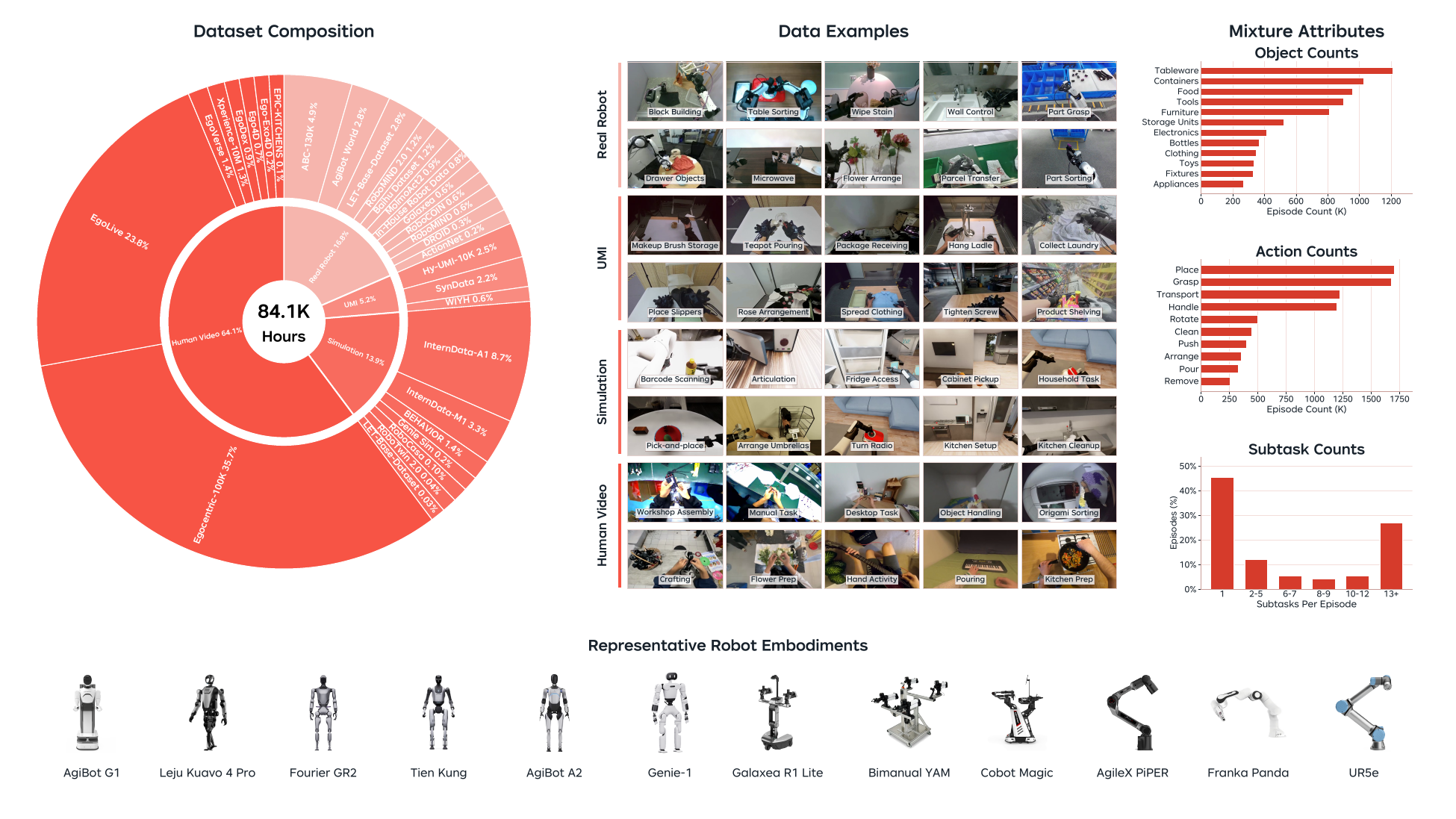}
    \captionof{figure}{\textbf{Overview of the 84.1K-hour training corpus.} Left:
    composition by data family and dataset. Center: representative observations
    from real-robot, UMI, simulation, and human-video data. Right: object,
    action, and subtask statistics. Bottom: representative robot embodiments.}
    \label{fig:dataset_overview}
\end{center}

\paragraph{Human egocentric videos.}
We collect human egocentric videos covering diverse objects, scenes,
and interaction patterns. The collection includes
EgoLive~\citep{li2026egolive},
Egocentric-100K~\citep{buildai2025egocentric100k},
EgoVerse~\citep{punamiya2026egoverse},
Xperience-10M~\citep{ropedia2026xperience10m},
Ego4D~\citep{grauman2022ego4d},
Ego-Exo4D~\citep{grauman2024egoexo4d}, EgoDex~\citep{hoque2025egodex} and
EPIC-KITCHENS-100~\citep{damen2022epickitchens100}. We filter and subsample the
public sources and process a subset of EgoLive using the JoyAI-Sim
pipeline~\citep{liu2026joyaisim}. These data broaden the diversity of manipulation
behaviors and provide large-scale visual-transition supervision for WLAM.

\paragraph{Real-robot trajectories.}
We combine public and internally collected robot trajectories spanning single-arm,
bimanual, mobile-manipulation, and humanoid platforms. Representative sources
include AgiBot World~\citep{bu2025agibotworld},
ABC-130K~\citep{allshire2026abc}, the Galaxea Open-World
Dataset~\citep{jiang2025galaxeaopenworld}, RoboCOIN~\citep{wu2025robocoin}, and
RoboMIND~\mbox{\citep{wu2025robomind}}, together with our internally collected AgiBot G1
multi-task data. These trajectories ground the learned transition representation
in physical robot interactions across diverse embodiments.

\begin{center}
    \begin{minipage}{\linewidth}
    \centering
    \scriptsize
    \fontfamily{ptm}\selectfont
    \setlength{\tabcolsep}{2pt}
    \renewcommand{\arraystretch}{0.96}
    \captionof{table}
    [Dataset composition and use across training stages]
    {\textbf{Dataset composition and use across training stages.}
    Size is measured in hours after filtering and subsampling.
    Markers \datasetstageWLAMmarker\ ,
    \datasetstageLACWMmarker\ , and
    \datasetstageGPMmarker\ denote WLAM training,
    LAC-WM pre-training, and generalist policy model mid-training,
    respectively.}
    \label{tab:lam_data}

    \begin{tabularx}{\linewidth}{@{}l l c >{\raggedright\arraybackslash}X l@{}}
        \toprule
        Data Family
        & Dataset
        & Size (Hours)
        & Embodiment
        & Training Stage \\
        \midrule

        \multirow[c]{12}{*}[-13pt]{Real Robot}
        & \datasetname{ABC-130K~\citeyearpar{allshire2026abc}}
        & 4,131
        & Bimanual YAM
        & \datasetstageall \\

        & \datasetname{AgiBot World~\citeyearpar{bu2025agibotworld}}
        & 2,392
        & AgiBot G1
        & \datasetstageall \\

        & \datasetname{LET-Base-Dataset~\textnormal{(Real Robot)}~\citeyearpar{leju2025let}}
        & 2,320
        & Leju Kuavo 4 Pro
        & \datasetstageWLAMmarker \\

        & \datasetname{RoboMIND 2.0~\citeyearpar{hou2025robomind2}}
        & 1,000
        & Franka Panda, UR5e, Tien Kung, etc.
        & \datasetstageWLAMmarker \\

        & \datasetname{Baihu Dataset~\citeyearpar{openloong2026baihu}}
        & 1,000
        & Qinglong, Fourier GR2, AgiBot A2, etc.
        & \datasetstageWLAMLACWM \\

        & \datasetname{MolmoAct2-BimanualYAM~\citeyearpar{fang2026molmoact2}}
        & 720
        & Bimanual YAM
        & \datasetstageWLAMLACWM \\

        & \datasetname{In-House Robot Data}
        & 675
        & AgiBot G1
        & \datasetstageall \\

        & \datasetname{Galaxea Open-World Dataset~\citeyearpar{jiang2025galaxeaopenworld}}
        & 500
        & Galaxea R1 Lite
        & \datasetstageall \\

        & \datasetname{RoboCOIN~\citeyearpar{wu2025robocoin}}
        & 500
        & Cobot Magic, RMC-AIDA-L, G1edu-u3, etc.
        & \datasetstageWLAMLACWM \\

        & \datasetname{RoboMIND~\citeyearpar{wu2025robomind}}
        & 500
        & Franka Panda, UR5e, Cobot Magic, etc.
        & \datasetstageWLAMmarker \\

        & \datasetname{DROID~\citeyearpar{khazatsky2024droid}}
        & 273
        & Franka Panda
        & \datasetstageWLAMmarker \\

        & \datasetname{ActionNet~\citeyearpar{fourier2025actionnet}}
        & 140
        & Fourier GR1, Fourier GR2
        & \datasetstageWLAMLACWM \\
        \midrule

        \multirow[c]{3}{*}[-2pt]{UMI}
        & \datasetname{Hy-UMI-10K~\citeyearpar{zhang2026hyembodied}}
        & 2,110
        & Custom-Designed Hand-Held UMI Grippers
        & \datasetstageWLAMGPM \\

        & \datasetname{SynData~\citeyearpar{psibot2026syndata}}
        & 1,818
        & PsiBot Exoskeleton Gloves
        & \datasetstageWLAMmarker \\

        & \datasetname{WIYH~\citeyearpar{zheng2025wiyh}}
        & 465
        & Oracle Suite H-Gloves
        & \datasetstageWLAMmarker \\
        \midrule

        \multirow[c]{7}{*}[-7pt]{Simulation}
        & \datasetname{InternData-A1~\citeyearpar{tian2025interndataa1}}
        & 7,330
        & Genie-1, Franka Panda, Split ALOHA, etc.
        & \datasetstageall \\

        & \datasetname{InternData-M1~\citeyearpar{internrobotics2025interndatam1}}
        & 2,800
        & Franka Panda, Cobot Magic, AgileX PiPER
        & \datasetstageWLAMmarker \\

        & \datasetname{BEHAVIOR-1K~\citeyearpar{li2023behavior1k}}
        & 1,200
        & Galaxea R1 Pro
        & \datasetstageWLAMLACWM \\

        & \datasetname{Genie Sim 3.0~\citeyearpar{yin2026geniesim3}}
        & 200
        & AgiBot G1, AgiBot G2
        & \datasetstageWLAMLACWM \\

        & \datasetname{RoboCasa~\citeyearpar{nasiriany2024robocasa}}
        & 84
        & Fourier GR-1
        & \datasetstageWLAMmarker \\

        & \datasetname{RoboTwin 2.0~\citeyearpar{chen2025robotwin}}
        & 34
        & ALOHA-AgileX
        & \datasetstageWLAMLACWM \\

        & \datasetname{LET-Base-Dataset~\textnormal{(Simulation)}~\citeyearpar{leju2025let}}
        & 28
        & Leju Kuavo 4 Pro
        & \datasetstageWLAMmarker \\
        \midrule

        \multirow[c]{8}{*}[-8pt]{Human Video}
        & \datasetname{Egocentric-100K~\citeyearpar{buildai2025egocentric100k}}
        & 30,000
        & \multirow[c]{8}{=}[-8pt]{Human}
        & \datasetstageWLAMmarker \\

        & \datasetname{EgoLive~\citeyearpar{li2026egolive}}
        & 20,000
        &
        & \datasetstageall \\

        & \datasetname{EgoVerse~\citeyearpar{punamiya2026egoverse}}
        & 1,200
        &
        & \datasetstageall \\

        & \datasetname{Xperience-10M~\citeyearpar{ropedia2026xperience10m}}
        & 1,059
        &
        & \datasetstageall \\

        & \datasetname{EgoDex~\citeyearpar{hoque2025egodex}}
        & 715
        &
        & \datasetstageall \\

        & \datasetname{Ego4D~\citeyearpar{grauman2022ego4d}}
        & 600
        &
        & \datasetstageWLAMLACWM \\

        & \datasetname{Ego-Exo4D~\citeyearpar{grauman2024egoexo4d}}
        & 200
        &
        & \datasetstageWLAMmarker \\

        & \datasetname{EPIC-KITCHENS-100~\citeyearpar{damen2022epickitchens100}}
        & 100
        &
        & \datasetstageWLAMmarker \\
        \bottomrule
    \end{tabularx}
    \end{minipage}
\end{center}

\paragraph{Simulation trajectories.}
We use simulation trajectories from InternData-A1~\citep{tian2025interndataa1},
InternData-M1~\citep{internrobotics2025interndatam1},
BEHAVIOR-1K~\citep{li2023behavior1k}, Genie Sim
3.0~\citep{yin2026geniesim3}, RoboCasa~\citep{nasiriany2024robocasa}, RoboTwin
2.0~\citep{chen2025robotwin}, and LET-Base-Dataset~\citep{leju2025let}. These data
further extend the coverage of the overall training corpus across scenes, objects,
tasks, and robot embodiments.

\paragraph{UMI demonstrations.}
The UMI data comprise Hy-UMI-10K~\citep{zhang2026hyembodied},
SynData~\citep{psibot2026syndata}, and WIYH~\citep{zheng2025wiyh}. These sources
span custom-designed hand-held UMI grippers and wearable glove interfaces,
providing human-centered manipulation demonstrations with synchronized visual
observations and interface motion. They complement robot trajectories with
diverse close-range interactions and portable data-collection settings.

\section{Experiments}
\label{sec:experiments}

\subsection{Experiment Setup}
\label{sec:exp_setup}

\begin{figure}[!tbp]
    \centering
    \includegraphics[width=\linewidth]{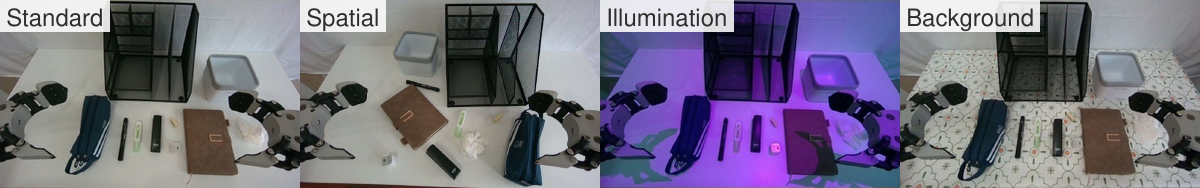}
    \caption{\textbf{Representative real-world evaluation scenes.} From left to right:
    the standard configuration (Standard), the spatial generalization setting
    with perturbed object placements and relative spatial relations (Spatial),
    the illumination generalization setting under altered ambient lighting
    (Illumination), and the background generalization setting with an unseen
    tablecloth texture (Background).}
    \label{fig:real_world_generalization_scenes}
\end{figure}

\paragraph{Real-World Benchmark.}
We evaluate the generalist policy on six tabletop manipulation tasks using
the AgiBot G1 platform after generalist policy model post-training. The six
tasks are to place an eraser on a shelf, place a charger on a shelf, place a
pencil case on a shelf, place a stapler on a shelf, move books to a bookshelf,
and discard trash in a bin. For brevity, we refer to these tasks as
\emph{Eraser}, \emph{Charger}, \emph{Pencil Case}, \emph{Stapler}, \emph{Book},
and \emph{Trash}, respectively. Together, these tasks assess fine-grained
object grasping and placement with both the left and right arms. For each task,
we evaluate three unseen generalization settings, as Figure~\ref{fig:real_world_generalization_scenes} shows. Spatial generalization perturbs
the initial position of the target object or alters its spatial relationship with
other task-relevant objects. Illumination generalization changes lighting, while background generalization replaces the tabletop covering
with unseen textures. Across all settings, the instruction and success criterion
remain unchanged.

\subsection{Main Results}
\label{sec:main_results}

\subsubsection{Latent-Action Representation on LARYBench}
\label{sec:lary_results}

We evaluate frozen representations with LARYBench~\citep{nie2026lary} using
semantic-classification and low-level control-regression probes. Following its
representation-level protocol, we evaluate the 1024D probing feature for
continuous LAMs and the pre-quantization feature for discrete LAMs; we evaluate the final latent for models whose final representations are available. We compare against
DiLA~\citep{zhang2026dila}, MVP-LAM~\citep{lee2026mvplam},
CLAP~\citep{zhang2026clap}, and
DreamDojo~\citep{gao2026dreamdojo}.
Table~\ref{tab:lary_standard} additionally incorporates a subset of baseline results reported by LARYBench, as indicated by the dagger; all unmarked results are evaluated by us. The standard protocol measures the information available in each model's representation, while the final-latent protocol tests how much of that information
is retained by the compact interface used downstream.

\begin{table}[!htbp]
    \centering
    \caption{\textbf{Standard representation evaluation on LARYBench.} Classification is
    top-1 accuracy (\%, $\uparrow$), and control regression is MSE ($\downarrow$).
    Best results are bold and second-best results are underlined. $^{\dagger}$
    denotes results quoted from LARYBench; all unmarked results are evaluated by
    us. Classification Avg. is the arithmetic mean over the two Composite
    datasets, and Regression Avg. is the arithmetic mean over the four control
    datasets.}
    \label{tab:lary_standard}
    \footnotesize
    \setlength{\tabcolsep}{1.5pt}
    \renewcommand{\arraystretch}{1.08}
    \begin{tabular*}{\textwidth}{@{\extracolsep{\fill}}lccc@{\hspace{5pt}}ccccc@{}}
        \toprule
        \multicolumn{1}{c}{\multirow{2}{*}{{\fontsize{9.5}{10.5}\selectfont\bfseries Method}}} &
        \multicolumn{3}{c}{{\fontsize{9.5}{10.5}\selectfont\bfseries Semantic classification $\uparrow$}} &
        \multicolumn{5}{c}{{\fontsize{9.5}{10.5}\selectfont\bfseries Control regression (MSE) $\downarrow$}} \\
        \cmidrule(lr){2-4}\cmidrule(lr){5-9}
        & \makecell{Composite\\Human}
        & \makecell{Composite\\Robot}
        & Avg.
        & CALVIN & VLABench
        & RoboCOIN & AgiBotBeta & Avg. \\
        \midrule
        LAPA$^{\dagger}$ & 14.61 & 23.64 & 19.13
        & 0.96 & 0.95 & 0.96 & 1.00 & 0.97 \\
        UniVLA$^{\dagger}$ & 19.08 & 18.56 & 18.82
        & 0.82 & 0.74 & 0.94 & 0.97 & 0.87 \\
        villa-X$^{\dagger}$ & 17.80 & 29.90 & 23.85
        & 0.86 & 0.72 & 0.94 & 0.97 & 0.87 \\
        DiLA & \underline{64.41} & 62.10 & \underline{63.26}
        & 0.55 & 0.27 & 0.86 & 0.82 & 0.63 \\
        MVP-LAM & 50.31 & 51.61 & 50.96
        & 0.55 & 0.38 & 0.90 & 0.92 & 0.69 \\
        CLAP & 55.32 & \underline{63.02} & 59.17
        & 0.43 & 0.17
        & \underline{0.68} & \underline{0.60} & \underline{0.47} \\
        DreamDojo & 63.69 & 59.41 & 61.55
        & \underline{0.35} & \underline{0.12}
        & 0.74 & 0.81 & 0.51 \\
        Ours & {\bfseries 73.12} & {\bfseries 68.37} & {\bfseries 70.75}
        & {\bfseries 0.25} & {\bfseries 0.09}
        & {\bfseries 0.55} & {\bfseries 0.57} & {\bfseries 0.37} \\
        \bottomrule
    \end{tabular*}
\end{table}

Table~\ref{tab:lary_standard} shows that our standard representation ranks first
on both Composite classification tasks and all four control datasets. Its 70.75\%
classification Avg. is 7.49 percentage points above DiLA's 63.26\%, while its regression
Avg. MSE of 0.37 is 21\% lower than CLAP's 0.47. Notably, neither CALVIN nor
VLABench is used in our LAM training, yet our representation reaches MSEs of
0.25 and 0.09, compared with DreamDojo's 0.35 and 0.12, respectively. This
demonstrates that the learned transition representation transfers
control-relevant structure beyond its training data.

Importantly, LARYBench exercises only our primary egocentric-view pathway because it does not
provide the hand-view or body-state inputs supported by our model.
Table~\ref{tab:lary_final} complements the standard protocol by evaluating the
representation actually passed to downstream models: intermediate and
pre-quantization features probe representational capacity, whereas the final
latent measures the information retained at the downstream interface. Despite a
$32\times$ compression from 1024D to 32D, our final latent retains comparable
classification performance, reaching 67.89\% versus 70.75\% for the standard
representation, and achieves the best average classification and regression
among final latent actions. This indicates that the compact interface retains
rich semantic and control-relevant information. Relative to their 1024D representations, the 32D latents of DreamDojo,
CLAP, and MVP-LAM reduce average classification accuracy by 9.16, 12.42,
and 32.62 percentage points, respectively. DiLA remains broadly stable,
but its classification accuracy still trails ours. AgiBotBeta's
stride-45 setting instead targets a different temporal regime,
asking one latent to span a much longer, potentially multi-stage transition than
the short transitions our latent is designed to represent.

\begin{table}[!htbp]
    \centering
    \caption{\textbf{Final latent-action evaluation on LARYBench.} Metrics and emphasis
    follow Table~\ref{tab:lary_standard}.}
    \label{tab:lary_final}
    \footnotesize
    \setlength{\tabcolsep}{1.5pt}
    \renewcommand{\arraystretch}{1.08}
    \begin{tabular*}{\textwidth}{@{\extracolsep{\fill}}lccc@{\hspace{5pt}}ccccc@{}}
        \toprule
        \multicolumn{1}{c}{\multirow{2}{*}{{\fontsize{9.5}{10.5}\selectfont\bfseries Method}}} &
        \multicolumn{3}{c}{{\fontsize{9.5}{10.5}\selectfont\bfseries Semantic classification $\uparrow$}} &
        \multicolumn{5}{c}{{\fontsize{9.5}{10.5}\selectfont\bfseries Control regression (MSE) $\downarrow$}} \\
        \cmidrule(lr){2-4}\cmidrule(lr){5-9}
        & \makecell{Composite\\Human}
        & \makecell{Composite\\Robot}
        & Avg.
        & CALVIN & VLABench
        & RoboCOIN & AgiBotBeta & Avg. \\
        \midrule
        DiLA & \underline{65.43} & \underline{62.50} & \underline{63.97}
        & 0.58 & 0.37 & 0.88 & 0.84 & 0.67 \\
        MVP-LAM & 16.78 & 19.89 & 18.34
        & 0.85 & 0.77 & 0.98 & 1.00 & 0.90 \\
        CLAP & 37.79 & 55.71 & 46.75
        & 0.64 & 0.31 & \underline{0.79} & {\bfseries 0.68} & \underline{0.61} \\
        DreamDojo & 54.22 & 50.55 & 52.39
        & \underline{0.47} & \underline{0.24} & 0.93 & 0.93 & 0.65 \\
        Ours & {\bfseries 69.21} & {\bfseries 66.56} & {\bfseries 67.89}
        & {\bfseries 0.33} & {\bfseries 0.12}
        & {\bfseries 0.75} & \underline{0.74} & {\bfseries 0.49} \\
        \bottomrule
    \end{tabular*}
\end{table}

\begin{figure}[!tbp]
    \centering
    \includegraphics[width=\textwidth]{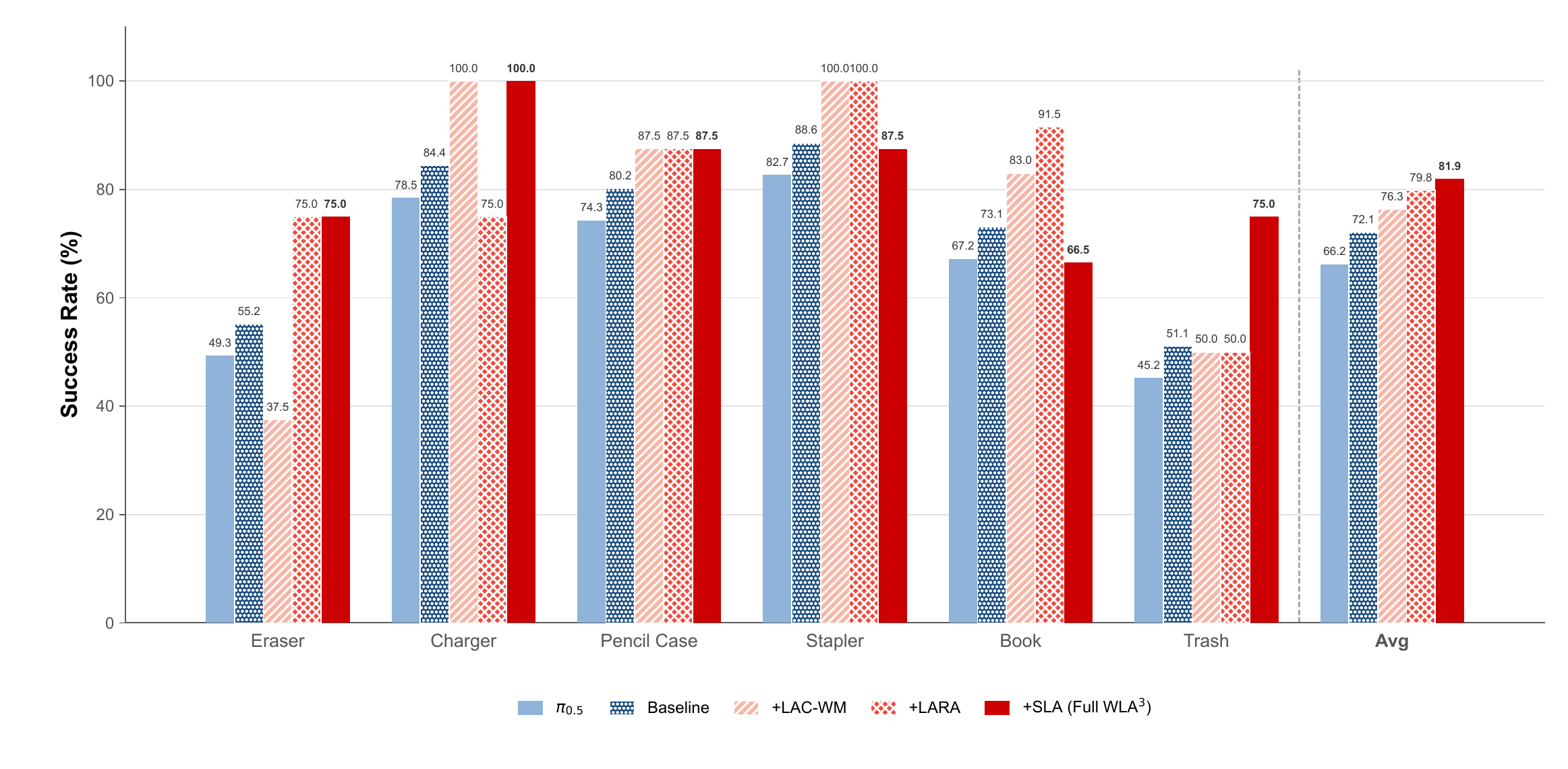}
    \caption{\textbf{Task-level real-world policy performance and cumulative
    component ablations.} Task success rate (SR, \%) is reported for six AgiBot
    G1 tabletop tasks under the unseen-generalization protocol described in
    Section~\ref{sec:exp_setup}. Each task score is averaged across the spatial,
    illumination, and background settings, and Avg. denotes the macro-average
    over the six tasks. The $\pi_{0.5}$ policy serves as the external comparator,
    whereas Baseline and the three component-addition variants form the internal
    cumulative ablation. Each successive variant retains the components
    introduced by the preceding variant, and +SLA denotes the full WLA$^3$
    model.}
    \label{fig:object_task_ablation}
\end{figure}

\subsubsection{Real-World Policy Evaluation}
\label{sec:real_world_policy_results}

We evaluate real-world policy performance on the six-task AgiBot G1 benchmark
described in Section~\ref{sec:exp_setup}. Figure~\ref{fig:object_task_ablation}
reports task-level success rates under the three unseen generalization settings,
together with the macro-average across tasks. All policies are post-trained on the same task-specific robot
demonstrations and evaluated under identical task instructions,
success criteria, and test configurations. These
perturbations test whether the policy maintains task execution as object placement,
illumination, and background appearance change.

The full WLA$^3$ model achieves an average success rate of 81.9\%, exceeding
$\pi_{0.5}$ by 15.7 percentage points. WLA$^3$ performs better on five of the
six tasks, with the largest gains on Trash, Eraser, and Charger, where it
improves by 29.8, 25.7, and 21.5 percentage points, respectively. On Book, its
success rate remains within 0.7 percentage points of $\pi_{0.5}$. These results
indicate that WLA$^3$ maintains stronger task execution under the evaluated
spatial, illumination, and background shifts.

\subsection{Scaling with Data}
\label{sec:scaling}

We study how policy performance scales with the amount of heterogeneous data
used for generalist policy model mid-training.
This experiment tests whether the shared transition representation continues to
convert additional heterogeneous data into useful policy supervision.

\begin{wrapfigure}{r}{0.48\textwidth}
    \centering
    \includegraphics[width=\linewidth]{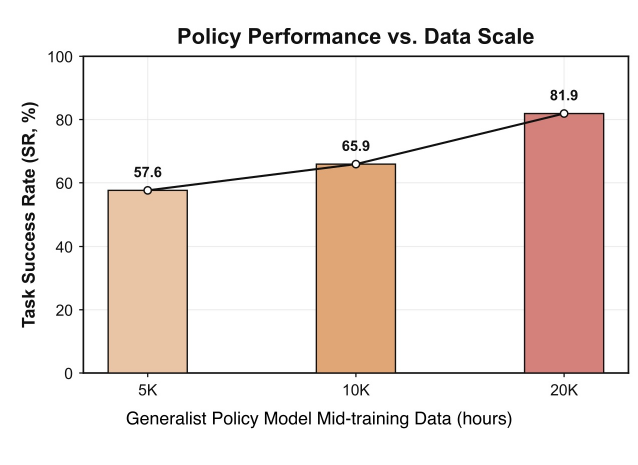}
    \caption{\textbf{Scaling generalist policy model mid-training data for WLA$^3$.}
    Generalist policy model post-training and evaluation settings are held fixed across data scales.}
    \label{fig:policy_data_scaling}
\end{wrapfigure}

From the same generalist policy model mid-training pool, we randomly construct subsets containing
5K, 10K, and 20K hours and train the same policy configuration
separately at each scale. All
variants use the same pretrained WLAM and LAC-WM checkpoints, model architecture,
initialization, and optimization recipe; only the amount of generalist policy model
mid-training data is varied. After generalist policy model mid-training, every model undergoes the same
generalist policy model post-training using the same task-specific dataset and training schedule. The
task-specific demonstrations are therefore held fixed and are not included in
the reported data volumes.

Each model after generalist policy model post-training is evaluated on the same six-task
real-world benchmark under three unseen generalization settings: spatial perturbations,
substantially altered ambient lighting, and unseen tabletop backgrounds. Across all
settings, the instruction and success criterion remain unchanged. We report the task
success rate averaged across tasks and generalization settings, as shown in
Figure~\ref{fig:policy_data_scaling}. The average success rate increases from
57.6\% with 5K hours to 65.9\% with 10K hours and 81.9\% with 20K hours. Since
the pretrained checkpoints, post-training data, and evaluation protocol are fixed,
the upward trend reflects the benefit of increasing the generalist policy model
mid-training mixture.

\subsection{Ablation Studies}
\label{sec:ablation}
\label{sec:lam_ablation}

\paragraph{Generalist Policy Ablation.} We conduct a cumulative component ablation under a shared policy-training and evaluation protocol. The baseline uses the VLM and a Perceiver-based Flow-Matching expert, initialized through generalist policy model mid-training and then adapted through generalist policy model post-training on robot demonstrations to predict 30-step chunks of 32D robot actions from visual features and the current robot state. The incremental variants add the three WLA$^3$ components in sequence. First, LAC-WM features are projected and concatenated with the VLM features, testing the contribution of the learned dynamics context. Next, LARA jointly predicts the 32D latent action and the native robot action at each step, preserving local transition supervision alongside executable control. Finally, SLA supervision adds a segment-level transition target for the VLM. All variants use the same data, initialization, and evaluation protocol, and each retains the components introduced by the preceding variant.

As shown in Figure~\ref{fig:object_task_ablation}, average success increases from 72.1\% for the baseline to 76.3\%, 79.8\%, and 81.9\% as LAC-WM, LARA, and SLA are introduced, respectively. The corresponding gains of 4.2, 3.5, and 2.1 percentage points show complementary benefits from dynamics context, latent-action prediction, and segment-level VLM supervision. The full model therefore combines improvements from all three components rather than relying on a single source of performance gain.

\paragraph{Latent Action Model Ablation.}
We retain five necessary variants and evaluate their final 32D latent
actions on LARYBench. Primary view only uses the primary camera transition.
Primary + wrist views adds synchronized wrist or hand views while omitting
embodiment-state input; comparing these two variants measures the contribution of
the additional viewpoints, while comparison with Full WLAM measures the contribution
of embodiment state. The two removal variants start from Full WLAM and disable only
cross-modal transition dropout or temporal overlap consistency, respectively. All
variants share training data, latent dimension, optimization settings, and
LARYBench protocol.
\begin{table*}[!htbp]
    \centering
    \caption{\textbf{WLAM ablations on LARYBench using the 32D latent
    action.} Classification is top-1 accuracy (\%, $\uparrow$), and control
    regression is MSE ($\downarrow$). Metrics and emphasis follow Table 2.}
    \label{tab:wlam_ablation}
    \scriptsize
    \setlength{\tabcolsep}{1.5pt}
    \renewcommand{\arraystretch}{1.08}
    \begin{tabular*}{\textwidth}{@{\extracolsep{\fill}}lccc@{\hspace{5pt}}ccccc@{}}
        \toprule
        \multicolumn{1}{c}{\multirow{2}{*}{Variant}} &
        \multicolumn{3}{c}{Semantic classification $\uparrow$} &
        \multicolumn{5}{c}{Control regression (MSE) $\downarrow$} \\
        \cmidrule(lr){2-4}\cmidrule(lr){5-9}
        & \makecell{Composite\\Human}
        & \makecell{Composite\\Robot}
        & Avg.
        & CALVIN & VLABench & RoboCOIN & AgiBotBeta & Avg. \\
        \midrule
        Primary view only
        & 61.85 & 59.23 & 60.54 & 0.49 & 0.25 & 0.86 & 0.85 & 0.61 \\
        Primary + wrist views (no state)
        & 66.12 & 63.50 & 64.81 & 0.41 & 0.18 & 0.81 & 0.79 & 0.55 \\
        Full w/o cross-modal dropout
        & 62.18 & 58.74 & 60.46 & 0.48 & 0.23 & 0.85 & 0.83 & 0.60 \\
        Full w/o overlap consistency
        & \underline{67.45} & \underline{64.81} & \underline{66.13} & \underline{0.37} & \underline{0.15} & \underline{0.78} & \underline{0.77} & \underline{0.52} \\
        Full WLAM
        & {\bfseries 69.21} & {\bfseries 66.56} & {\bfseries 67.89}
        & {\bfseries 0.33} & {\bfseries 0.12}
        & {\bfseries 0.75} & {\bfseries 0.74} & {\bfseries 0.49} \\
        \bottomrule
    \end{tabular*}
\end{table*}

As shown in Table~\ref{tab:wlam_ablation}, adding wrist views improves the average classification accuracy from 60.54\% to
64.81\% and reduces the average control-regression MSE from 0.61 to 0.55.
Incorporating embodiment state further raises classification accuracy to 67.89\%
and lowers MSE to 0.49. Removing cross-modal transition dropout causes the
largest degradation, reducing classification accuracy to 60.46\% and increasing
MSE to 0.60, while removing overlap consistency yields 66.13\% and 0.52.
Together, these results show that multi-view observations and embodiment state
provide complementary transition evidence, while both regularizers improve the
quality of the final compact latent action.

\subsection{Human-Robot Alignment Experiment}
\label{sec:human_video_post_training}

To assess the contribution of task-matched human demonstrations under
different robot-data regimes, we compare three post-training settings.
\textbf{Full-Robot} uses the full original set of Eraser robot
demonstrations and no Eraser human demonstrations.
\textbf{Limited-Robot} retains only one twentieth (5\%) of the Eraser
robot demonstrations and includes no Eraser human demonstrations.
\textbf{Limited-Robot+Human} uses the same reduced robot subset as
Limited-Robot and additionally incorporates task-matched Eraser human
demonstrations. During generalist policy model post-training, robot trajectories
supervise both latent-action and robot-action prediction. Human
demonstrations provide only latent-action supervision because they do
not contain executable robot-action labels; their robot-action loss is
therefore masked. Across all three settings, the initialization, data for
the other five tasks, and all other post-training configurations are
kept unchanged.

\begin{table*}[!htbp]
    \centering
    \caption{\textbf{Effect of task-matched human demonstrations across Eraser robot-data regimes.} All settings use identical initialization, data for
    the other five tasks, and post-training configurations. SR denotes
    success rate (\%, $\uparrow$).}
    \label{tab:human_robot_alignment}
    \small
    \begin{tabular}{lccc}
        \toprule
        \makecell{Post-training setting}
        & \makecell{Eraser robot data}
        & \makecell{Eraser human data}
        & \makecell{Eraser SR (\%) $\uparrow$} \\
        \midrule
        Full-Robot          & 100\% & No  & 75 \\
        Limited-Robot       & 5\%   & No  & 40 \\
        Limited-Robot+Human & 5\%   & Yes & 60 \\
        \bottomrule
    \end{tabular}
\end{table*}

As shown in Table~\ref{tab:human_robot_alignment}, reducing the Eraser
robot data from 100\% to 5\% decreases the success rate from 75\% to
40\%. Adding task-matched human demonstrations under the same limited
robot-data regime improves the success rate from 40\% to 60\%, an
absolute gain of 20 percentage points. This recovers 20 of the
35 percentage points lost after reducing the robot data, while remaining
15 percentage points below the Full-Robot setting. These results suggest
that task-matched human demonstrations provide useful task-relevant
signals through latent-action supervision and can partially compensate
for limited robot demonstrations, although they do not fully replace
the benefit of executable robot-action supervision in this experiment.

\begin{figure}[!tbp]
    \centering
    \includegraphics[width=\textwidth]{
        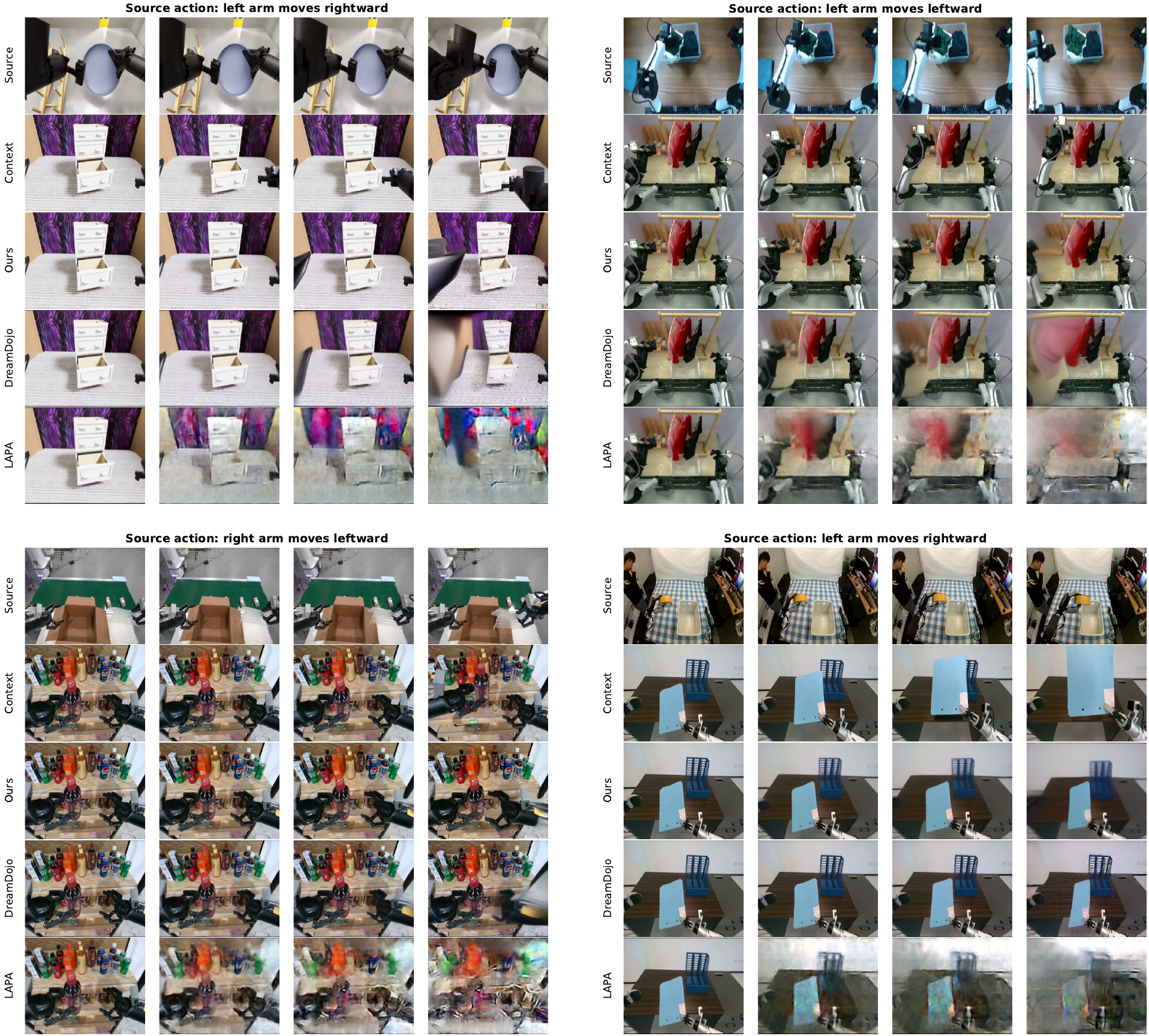
    }
    \caption{\textbf{Cross-context transition transfer.}
    Four source--context pairs compare WLAM, DreamDojo, and LAPA in multi-step
    autoregressive rollouts. Panel titles describe the source transition, and the
    four columns show frames sampled throughout each rollout. Source and Context
    are ground-truth trajectories. Each model starts from the first context frame
    and predicts subsequent frames using latent actions from the source.}
    \label{fig:cross_model_latent_swap}
\end{figure}

\subsection{Representation Analysis}
\label{sec:analysis}


\paragraph{Cross-Embodiment Representation Transfer.}
We evaluate the latent-action representation under both within-domain and
cross-embodiment settings. The Composite Human and Composite Robot results in
Table~\ref{tab:lary_final} measure whether action information can be decoded
within each domain, using probes trained and evaluated on human or robot data,
respectively. To assess cross-embodiment transfer, we select twelve action
classes shared by human and robot videos. For all methods, representations are
mapped by the same frozen adapter to four 32D tokens and evaluated using the
same linear probe. H$\rightarrow$H and R$\rightarrow$R denote within-domain
evaluation, whereas H$\rightarrow$R and R$\rightarrow$H denote transfer from
human to robot and from robot to human, respectively.

As shown in Table~\ref{tab:cross_embodiment_transfer}, WLAM achieves the highest
accuracy in all four directions. Its average accuracy across the two
cross-embodiment directions reaches 20.98\%, exceeding DiLA by 5.27 percentage
points. Under the source-family-disjoint split and shared evaluation protocol,
these results indicate that WLAM captures action information that remains
readable across human and robot embodiments. Together with the LARYBench
results, they demonstrate that WLAM produces informative representations
within each domain and transfers more effectively across embodiments.

\begin{table}[!htbp]
    \centering
    \caption{\textbf{Cross-embodiment representation transfer.} Entries are top-1
    accuracy (\%, $\uparrow$) from the same linear reader and fixed 4$\times$32
    token interface. H and R denote human and robot videos, respectively. Best results are bold and second-best results are underlined.}
    \label{tab:cross_embodiment_transfer}
    \footnotesize
    \setlength{\tabcolsep}{4pt}
    \renewcommand{\arraystretch}{1.08}
    \begin{tabular*}{\textwidth}{@{\extracolsep{\fill}}lccc@{\hspace{8pt}}ccc@{}}
        \toprule
        \multirow{2}{*}{Method} &
        \multicolumn{3}{c}{Within-domain} &
        \multicolumn{3}{c}{Cross-embodiment} \\
        \cmidrule(lr){2-4}\cmidrule(lr){5-7}
        & H$\rightarrow$H & R$\rightarrow$R & Avg.
        & H$\rightarrow$R & R$\rightarrow$H & Avg. \\
        \midrule
        DiLA       & \underline{40.75} & 39.00 & \underline{39.88} & \underline{16.17} & \underline{15.25} & \underline{15.71} \\
        MVP-LAM    & 27.67 & 25.96 & 26.82 & 12.54 & 13.75 & 13.15 \\
        CLAP   & 28.88 & \underline{40.38} & 34.63 & 10.50 & 10.88 & 10.69 \\
        LAPA       & 22.21 & 17.33 & 19.77 & 14.21 & 14.42 & 14.31 \\
        UniVLA     & 29.25 & 27.00 & 28.13 & 12.21 & 13.38 & 12.79 \\
        villa-X    & 24.13 & 26.04 & 25.09 &  9.83 & 12.29 & 11.06 \\
        WLAM (ours)& {\bfseries 49.38} & {\bfseries 47.38} & {\bfseries 48.38}
                   & {\bfseries 22.08} & {\bfseries 19.88} & {\bfseries 20.98} \\
        \bottomrule
    \end{tabular*}
\end{table}

\paragraph{Cross-Context Transition Transfer.} A transition code should describe how a scene changes without being tied to
the appearance of its source trajectory. For each of four source--context
pairs, we extract latent actions from the source, initialize generation with
the first context frame, and predict the remaining frames autoregressively. We
apply the same procedure to WLAM, DreamDojo~\citep{gao2026dreamdojo}, and
LAPA~\citep{ye2024latent}, using each method's own representation,
conditioning inputs, and decoder. This experiment provides a qualitative
comparison rather than an aggregate score.

The examples in Figure~\ref{fig:cross_model_latent_swap} cover left- and
right-arm transitions in simulation and real-world scenes. WLAM follows the
source transition while retaining the objects and layout of the context.
DreamDojo also preserves the context but develops local blur or appearance
drift in later frames, whereas LAPA shows stronger texture and structural
degradation. These examples indicate that WLAM can transfer the source change
while preserving the appearance and geometry of the new context.

Together, these analyses show that WLAM learns a transition-focused code that
transfers across embodiments and visual contexts.

\section{Conclusion, Limitations, and Future Work}
\label{sec:conclusion}

We presented WLA$^3$, a multi-granularity latent-action framework for
learning generalist robot policies from heterogeneous interaction data.
WLAM treats world-state transitions as a shared action interface and learns
compact representations from human videos, robot trajectories, simulation
rollouts, and UMI demonstrations. WLA$^3$ uses these representations at
three levels: LAC-WM models action-conditioned dynamics, SLA provides
segment-level supervision to the VLM, and LARA jointly predicts shared
latent actions and embodiment-specific robot controls.

On the six-task real-world benchmark, WLA$^3$ achieves an average success
rate of 81.9\%, compared with 66.2\% for $\pi_{0.5}$. The cumulative
ablation improves the shared baseline from 72.1\% to 76.3\% with LAC-WM,
79.8\% with LARA, and 81.9\% with SLA. With only 5\% of the Eraser robot
demonstrations, task-matched human demonstrations improve success from
40\% to 60\%, partially closing the gap to the 75\% achieved with the full
robot dataset. On LARYBench, the standard WLAM representation achieves
70.75\% average classification accuracy and an average control-regression
MSE of 0.37. Its compact 32D latent retains 67.89\% classification accuracy
and achieves an MSE of 0.49, outperforming the other evaluated final latent
representations on both metrics.

The current world-state representation relies primarily on multi-view RGB
observations and robot proprioception, and therefore captures contact and
physical state only indirectly. Future work could incorporate measured
depth, tactile and force feedback, richer object-state information, and
high-confidence human hand trajectories. Broader language supervision and
more diverse training data may further improve fine-grained semantic
grounding and enable more systematic studies of scaling and transfer across
data sources and embodiments.

\bibliographystyle{conference}
\bibliography{paper}

\newpage
\appendix

\section{Additional WLAM Formulation Details}
\label{app:wlam_details}

\paragraph{Input Tokens and Posterior Parameterization.} For each transition, the endpoint observations and available embodiment states are
mapped to modality-specific tokens. A shared embodiment-profile token identifies the
source configuration. The transition encoder processes these tokens with a learned
$[\mathrm{CLS}]$ token, whose output forms $\mathbf{g}_{t,h}$. Only this resulting
$[\mathrm{CLS}]$ output is retained for posterior parameterization and construction
of the segment-level target. Two projection heads
predict the mean $\boldsymbol{\mu}_{t,h}$ and diagonal standard deviation
$\boldsymbol{\sigma}_{t,h}$ of the Gaussian posterior. Reconstruction uses the
reparameterized sample
\begin{equation}
    \mathbf{z}_{t,h}=\boldsymbol{\mu}_{t,h}
    +\boldsymbol{\sigma}_{t,h}\odot\boldsymbol{\epsilon},
    \qquad \boldsymbol{\epsilon}\sim\mathcal{N}(0,I_{d_z}),
\end{equation}
whereas consistency regularization and offline annotation use the posterior mean
$\bar{\mathbf{z}}_{t,h}=\boldsymbol{\mu}_{t,h}$.

\paragraph{Cross-Modal Transition Dropout.} The dropout mask is sampled over the camera and embodiment-state streams recorded for
each example, with at least one transition stream retained. The same mask is applied
at both endpoints so that masking does not introduce an artificial change. The
transition encoder receives the masked endpoint pair, whereas the decoder receives
the recorded initial-state modalities and reconstructs every recorded endpoint target.
Modalities absent from the source data are excluded from both the encoder input and the
corresponding reconstruction terms.
Offline targets are extracted from complete recorded transitions without transition
dropout.

\paragraph{Reconstruction and Temporal Sampling.} The reconstruction objective is
\begin{equation}
\begin{aligned}
    \mathcal{L}_{\mathrm{rec}}={}&
    \lambda_{\mathrm{pix}}\mathcal{L}_{\mathrm{pix}}
    +\lambda_{\mathrm{perc}}\mathcal{L}_{\mathrm{LPIPS}}
    +\lambda_{\mathrm{flow}}\mathcal{L}_{\mathrm{flow}}
    +\lambda_{\mathrm{DINO}}\mathcal{L}_{\mathrm{DINO}}\\
    &+\lambda_{\mathrm{depth}}\mathcal{L}_{\mathrm{depth}}
    +\lambda_{\mathrm{3D}}\mathcal{L}_{\mathrm{VGGT}}
    +\lambda_{\mathrm{state}}\mathcal{L}_{\mathrm{state}}.
\end{aligned}
\end{equation}
The visual terms use pixel and LPIPS reconstruction~\citep{zhang2018unreasonable},
RAFT optical-flow features~\citep{teed2020raft}, DINOv2 semantic
features~\citep{oquab2024dinov2}, Depth Anything V2~\citep{yang2024depthanythingv2},
and VGGT geometric features~\citep{wang2025vggt}. State-bearing examples additionally
reconstruct the recorded native embodiment state. Each example contributes only the
terms supported by its targets.

For temporal overlap consistency, we sample two windows with the same endpoint offset
$h$ and shift their starting indices by $\delta$, where $h>\delta>0$. The windows
therefore share $h-\delta$ interaction steps. Consistency is evaluated on posterior
means rather than stochastic samples.

\begingroup
\small
\setlength{\parskip}{2pt plus 1pt minus 1pt}
\setlength{\abovedisplayskip}{4pt plus 1pt minus 1pt}
\setlength{\belowdisplayskip}{4pt plus 1pt minus 1pt}
\section{Generalist Policy Training Details}
\label{app:policy_details}

\paragraph{State and Action Interfaces.}
For WLAM, $r_t$ denotes the native embodiment state when it is recorded; visual-only
trajectories omit this modality and its corresponding reconstruction target. For policy training, each robot's native state vector is zero-padded to 32 dimensions and used as the condition \(\mathbf{r}_t^{\mathrm{pol}}\), while human-video samples use an all-zero condition. The fixed robot-action interface and coordinate mask are
defined in Eq.~\ref{eq:policy_robot_action_interface}. At deployment, only valid
robot-action coordinates are interpreted in the target embodiment's native format.

\paragraph{LAC-WM Flow Matching.}
The flow path, velocity prediction, and LAC-WM pre-training objective are given in
Eqs.~\ref{eq:policy_lacwm_flow_path}--\ref{eq:policy_lacwm_objective}. At temporal
position $i$, the finite difference is
$\Delta_{\mathrm{pos}}\mathbf{v}_{\tau,i}
=\mathbf{v}_{\tau,i+1}-\mathbf{v}_{\tau,i}$. With observations always conditioned on, language and latent-action conditions are independently included or omitted, yielding four training modes: observation-only, language-conditioned, latent-action-conditioned, and jointly conditioned on both.

\paragraph{Masked Branch Losses and EMA Normalization.}
For training sample $n$ and branch $b\in\{\mathrm{lat},\mathrm{rob}\}$, the masked
Flow-Matching loss is
\begin{equation}
    \mathcal{L}_b=
    \frac{\sum_n\left\|\mathbf{M}_n^b\odot
    \left(\widehat{\mathbf{V}}_n^b-\mathbf{V}_n^{\star,b}\right)\right\|_F^2}
    {\sum_n\left\|\mathbf{M}_n^b\right\|_1+\varepsilon_{\mathrm{mask}}},
    \qquad b\in\{\mathrm{lat},\mathrm{rob}\}.
    \label{eq:policy_branch_loss}
\end{equation}
The mask selects the targets available in each trajectory and, for robot actions, the
valid native coordinates. Latent-only trajectories contribute to the latent branch,
whereas trajectories with both targets, including UMI data, contribute to both branches.
To balance their scales, we maintain a separate exponential moving average for each
branch loss:
\begin{equation}
    m_b\leftarrow\rho m_b+(1-\rho)\mathcal{L}_b,\qquad
    \widetilde{\mathcal{L}}_b=
    \frac{\mathcal{L}_b}{\operatorname{sg}(m_b)+\varepsilon_{\mathrm{ema}}},
    \quad b\in\{\mathrm{lat},\mathrm{rob}\},
    \label{eq:policy_ema_normalization}
\end{equation}
where $\operatorname{sg}$ stops gradients through the running statistic. These
normalized losses enter the generalist policy model mid-training objective in
Eq.~\ref{eq:policy_objective_main}. Numerical integration produces latent-action and
robot-action predictions; only the valid robot-action coordinates are executed during
deployment.
\endgroup

\end{document}